\documentclass[conference,final]{IEEEtran}
\IEEEoverridecommandlockouts
\usepackage{cite}
\usepackage{amsmath,amssymb,amsfonts}
\usepackage{graphicx}
\usepackage{textcomp}
\usepackage{xcolor}
\def\BibTeX{{\rm B\kern-.05em{\sc i\kern-.025em b}\kern-.08em
    T\kern-.1667em\lower.7ex\hbox{E}\kern-.125emX}}

\usepackage{tcolorbox}  
\usepackage{booktabs} 
\usepackage{tabularx}
\usepackage{graphics}
\usepackage{amsfonts}
\usepackage{epsfig}
\usepackage{url}
\usepackage{multirow}
\usepackage{xspace}
\usepackage{amssymb,wasysym,algorithm}
\usepackage[noend]{algpseudocode}
\usepackage{enumerate}
\usepackage{enumitem}
\usepackage{algcompatible}
\usepackage[bottom]{footmisc}
\usepackage{color}
\usepackage{soul}
\usepackage{mdwlist}
\usepackage{pgf}

\usepackage{booktabs} 
\usepackage{graphics}
\usepackage{amsmath}
\usepackage{amsfonts}
\usepackage{epsfig}
\usepackage{url}
\usepackage{multirow}
\usepackage{graphicx}
\usepackage{xspace}
\usepackage{amssymb,wasysym,algorithm}
\usepackage[noend]{algpseudocode}

\usepackage{enumitem}
\usepackage{algcompatible}
\usepackage[bottom]{footmisc}
\usepackage{color}
\usepackage{pdfpages}
\newtheorem{problem}{Problem}
\usepackage{mdwlist}
\usepackage{enumerate}
\usepackage{xcolor}
\usepackage{array}
\usepackage{colortbl}
\usepackage{wrapfig}
\usepackage{nohyperref} 

\newtheorem{lemma}{Lemma}

\newcommand{\Q}{\mathcal{Q}}
\newcommand{\s}{\mathcal{S}}

\newcommand{\qi}{\mathcal{Q}_I}
\newcommand{\qs}{\mathcal{Q}_c}
\newcommand{\gi}{\gamma_I}
\newcommand{\gs}{\gamma_c}
\newcommand{\sourceSet}{\mathcal{O}}

\newcommand{\given}[1][]{#1\vert}
\newcommand{\modelcost}{\textsc{ModelCost}\xspace} 
\newcommand{\algolocal}{\textsc{OnlyConnectivity}\xspace}
\newcommand{\jointalgolocal}{\textsc{JointPathMap}\xspace}

\newcommand{\prob}{\textsc{NetPathState}\xspace} 
 
\newcommand{\pathmap}{\textsc{OnlyConnectivity}\xspace} 
\newcommand{\norm}[1]{\lvert #1 \rvert}

\allowdisplaybreaks  

\begin{document}
\title{Mapping Network States using Connectivity Queries}

\author{Alexander Rodr\'iguez$^*$,  Bijaya Adhikari$^\dagger$, Andr\'es D. Gonz\'alez$^\circ$, Charles Nicholson$^\circ$, \\
Anil Vullikanti$^{\dagger\dagger}$ and B. Aditya Prakash$^*$\\
{\small $^*$College of Computing, Georgia Institute of Technology [\{arodriguezc, badityap\}@cc.gatech.edu]}\\
{\small $^\dagger$Department of Computer Science, University of Iowa [bijaya-adhikari@uiowa.edu]}\\
{\small $^\circ$School of Industrial and Systems Engineering, University of Oklahoma [\{andres.gonzalez, cnicholson\}@ou.edu]}\\
{\small $^{\dagger\dagger}$Biocomplexity Institute and Department of Computer Science, University of Virginia [vsakumar@virginia.edu]}
}


\maketitle

\vspace{-5cm}
\begin{abstract}
Can we infer all the failed components of an infrastructure network, given a sample of reachable nodes from supply nodes?
One of the most critical post-disruption processes after a natural disaster is to quickly determine the damage or failure states of critical infrastructure components. However, this is non-trivial, considering that often only a fraction of components may be accessible or observable after a disruptive event.  
Past work has looked into inferring failed components given point probes, i.e. with a direct sample of failed components. In contrast, we study the harder problem of inferring failed components given partial information of some `serviceable' reachable nodes and a small sample of point probes, being the first often more practical to obtain. 
We formulate this novel problem using the Minimum Description Length (MDL) principle, and then present a greedy algorithm that minimizes MDL cost effectively. We evaluate our algorithm on domain-expert simulations of real networks in the aftermath of an earthquake. Our algorithm successfully identifies failed components, especially the critical ones affecting the overall system performance.
\end{abstract}

\section{Introduction}
\label{sec:intro}

\noindent \textbf{Motivation.} 
Inferring network states from partial observations is an important problem in many applications, such as in epidemics, social networks and Internet/AS graphs. Consider critical infrastructure networks like water distribution system or electric grids --- their rapid functional restoration is a priority right after a disaster such as an earthquake~\cite{gomez_integrating_2019}. The sparse installation of real time monitoring systems installed in these networks (e.g. SCADA systems) imply that only a partial observation of the network is available post disaster i.e., we only get a sample of the functional states of the components (`failed' or not). However, for decision makers to devise an effective recovery plan, the information on functional states of the entire network is required. Hence, it is crucial to infer the states of the components in such networks after a disaster.

Despite its importance, there has been limited research in this space. Most past work has only looked into the so-called \emph{point queries}, i.e. one is given some direct sample of failed components. In epidemics, Shah et al.~\cite{shah2010detecting} proposed inferring missing infections and source detection given a sample of infections.
Indeed, in network tomography, all prior work has focused on independent failures~\cite{jin2006network}. 
Recently, there has been some progress on reconstructing network states assuming that the failures are geographically correlated given a sample of the failed components
~\cite{adhikari2018near}. As mentioned earlier, a common theme among these works is that the observed failures are a direct sample of failed components.

\begin{figure}[!t]
\small

\centering
\begin{tabular}{cc}
      \includegraphics[width=0.4\textwidth]{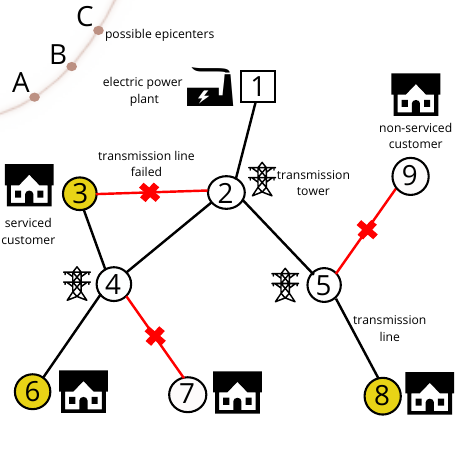} &  \\
      \vspace{5pt} (a) & \\
     \begin{tabular}[b]{|p{2.8cm}|p{4.0cm}|}
    \hline
    \textbf{Notation} & \textbf{Physical set in Fig. 1(a)} \\
    \hline
    Supply nodes $V_S$ & Electric power plant \\ & $= \{ 1\}$\\\hline
    Transshipment nodes $V_T$ &Transmission towers   $= \{ 2,4,5\}$\\\hline
    Demand nodes $V_D$& Houses/customers $= \{ 3,6,7,8,9\}$\\\hline
    Serviced nodes $\s$ & Houses with electricity \\ & $ = \{ 3,6,8\} \subseteq V_D $\\\hline
    True failure set $I$ & Transmission lines failed \\ & $= \{ (2,3), (4,7), (5,9)\}$\\\hline
    Disaster scenarios $\mathcal{O}$ & Possible epicenters $ = \{ A, B, C\}$ \\\hline
\end{tabular} 
\end{tabular}
(b)
\caption{
(a) An example of our problem using an electric power network disrupted by an earthquake. We refer to the demand nodes (houses) still connected to a supply node (power plant) as serviced nodes, i.e. houses with electricity. Thus, given a sample of serviced nodes $\{3,8\}$ (connectivity probes), a smaller sample of the set of failed edges $I$, and knowing that the epicenter can be A, B, or C, we want to infer the rest of true failure set (edges with crosses). 
Note that failed edge $\{2,3\}$ does not affect the serviceability of any node; therefore, using only connectivity probes will not allow us to discover this failure.
(b) Description of the physical sets in (a), and its association with our notation.
} 
\label{fig:teaser}
\end{figure}

A different class of queries, namely, \emph{connectivity queries} also occur naturally. For example,
consider an electric power grid, where transmission line connecting two component may fail~\cite{chen2017hotspots}. Such failures cause disconnections in the network.
SCADA systems (which oversee sensors in a power grid) provide information on
whether or not there is a power supply at a specific demand node (a consumer node). These demand nodes are supplied by identifiable supply nodes (power generators), 
thus, a stop of supply to a demand node can be mapped to disconnectivity 
between it and the supply nodes. Similarly, the continuous supply on a demand node indicates that there is at least one connected path to the supply nodes.
A motivating example is depicted in Figure~\ref{fig:teaser}. Here, node $1$ is a source node (power plant), the set $\{3,4,6,7,8,9\}$ is the set of demand nodes (consumers), and the remaining nodes $\{ 2,4,5\}$ are transmission nodes. If due to a disaster edges in the set $\{(2,3), (4,7), (5,9) \}$ fail, this would disconnect nodes $7$ and $9$.  

\linespread{1.25}

\noindent \textbf{Why study connectivity queries?} 
After a natural disaster, connectivity queries are \emph{more practical} to obtain than point queries. 
Information about supplied demand and customer service (a natural source of connectivity queries) is often relatively accessible---customers nowadays have multiple ways of reporting issues, including even social media \cite{mcclendon_leveraging_2013}---and has been proven to be useful to localize failures \cite{ji_resilience_2017}. In our example, a sample of the nodes connected to the source are observed i.e, we observe a subset of $\{3,6,8\}$ thanks to signals from sensors and/or consumer reports.

Motivated by the reasons above, we investigate the usefulness of adding connectivity queries to the point queries (collectively referred to as `joint queries') in inferring network failures. We find that using \emph{only} connectivity queries for inference is very challenging. For instance, in our example, no demand node is affected by the failure of edge $\{2,3\}$, which makes it undiscoverable to connectivity queries.
Hence, we propose using `joint queries' where we rely primarily on connectivity queries but also use a small sample of point queries to boost performance. To this end, we propose to the leverage the powerful Minimum Description Length (MDL)~\cite{grunwald04tutorial} concept, to find out both the number of failures and their identity. MDL is especially suited for this task, as it gives us a principled way to formulate our problem. 

\noindent \textbf{Contributions.} 
Our main contributions in this paper are as follows: 
\begin{enumerate}
\item \textit{Formulation and algorithm.} We formulate \prob{} to leverage joint probes for network state inference. We propose \jointalgolocal{} algorithm to solve the problem.
\item \textit{Extensive experiments on real datasets}. We use domain-expert simulations by civil engineers of three real network topologies in the aftermath of an earthquake. Our extensive experiments demonstrate that \jointalgolocal{} significantly outperforms the baselines.  
\end{enumerate}

\noindent To the best of our knowledge, our work is the first to propose connectivity queries and joint queries for network state inference.

\section{Preliminaries}  \label{sec:prelim}

\subsection{Context and Setup}
We are given an undirected graph $G(V,E)$ that represents an infrastructure network. Links in $E$ represent transmission systems and 
set $V$ contains three disjoint sets of nodes: supply nodes $V_S$, demand nodes $V_D$, and transshipment nodes $V_T$. Thus, $V=V_S \cup V_D \cup V_T$. Supply nodes are where flow (e.g. electricity, water) is originated, demand nodes are where flow is received and distributed to final users, and transshipment nodes are where flow is transferred to other link(s). 
We assume supply and demand nodes are related by many-to-many relationships (i.e. one supply node can service many demand nodes and one demand node can be serviced by many supply nodes) as it is observed in electric and water networks. This kind of structure appears in many different critical infrastructure networks \cite{chen2017hotspots}. 

In addition, we are given a set $\mathcal{O}$ containing disaster scenarios that characterize the most likely disaster hazards. For example, in case of earthquakes, they can be described by a magnitude and a geographical epicenter; even though there is a myriad of possibilities for this, typically there is a finite set of earthquake scenarios (our set $\mathcal{O}$) that can characterize most of the seismic hazard~\cite{adachi_comparative_2010}.  Once a scenario is characterized, the effects on the infrastructure network are driven by soil and structural properties in a stochastic manner.

\subsection{Failure Model} \label{sec:failure_model}
The failure model leverages `geographically correlated' failures, which have been extensively used in prior work both in the computer science~\cite{agarwal:infocom11} and civil engineering communities \cite{gomez_integrating_2019}. %
At an abstract level, given a disaster scenario $o \in \mathcal{O}$, some set of edges $I \subseteq E$ fail. 
Let $p(o)$ be the probability of a disaster scenario $o$ to had happen; hence, by previous definition, $\sum_{o\in\mathcal{O}} p(o) = 1$. 
Let $F(e|o)$ be the probability of edge $e$ to fail given $o$, e.g.  
probability of an electric transmission line to fail given an earthquake with epicenter $o$. 
These probability distributions are calculated considering not only the distance to the epicenter of $o$, but also the type of structure, construction materials, and condition.
In addition, we assume all edge failures depend only on the disaster's effect, i.e. failures are independent from each other. 
Both $\mathcal{O}$ and $F(e|o), \forall e\in E, \forall o\in \mathcal{O}$ are given by civil engineering guidelines, as in our experiments.

\section{Formulation and Approach}
\label{sec:problem_formulation}
\noindent Here we systematically formulate our problem based on network connectivity using the MDL principle. See Table~\ref{tab:notation} for notation. 

\subsection{Connectivity Probes} 
Once a failure has taken place in the network, an operator performs the so-called ``connectivity queries'' on the network, as described next, to gather some information on the status. 
Connectivity queries (coming from sensors and/or customer reports) serve as indirect observers for the network states.  As exemplified in Figure~\ref{fig:teaser}, when failures happen, they directly effect the \emph{serviceability} of some demand nodes in the network. Demand nodes require an active path to a nearby supply node to be serviced (e.g. receive electricity). 
Thus, due to failures because of the disaster, some demand nodes become non-serviced. 
We define a `sensor' set $\s \subseteq V_D$ as the set of demand nodes which are still serviced after removing failure edge set $I$ from $G$.  The operator will finally receive a \emph{connectivity probe} set $\Q_c \subseteq \s$. We can assume that it is chosen uniformly at random (with some given probability, say $\gs$), from $\s$. 
Therefore, we have the following dependency relationships:
\setlength\abovedisplayskip{5pt}
\setlength\belowdisplayskip{1pt}
\begin{equation}
o \rightarrow I \rightarrow \s \rightarrow \qs
\label{eq:causal}
\end{equation}

\subsection{Joint Probes} 
While connectivity queries are more practical to obtain, they are often not enough. If the demand nodes have multiple paths to the supply nodes (i.e., the network is redundant) the connectivity probes $Q_c$ fail to provide enough information to infer $I$ and $o$ correctly. We saw this in Figure~\ref{fig:teaser}, where failure of edge $\{2,3\}$ was not disconnecting any node, thus, it does not affect serviceability. 
It turns out that such phenomenon is quite common in real critical infrastructure networks as we empirically demonstrate it in Section~\ref{sec:expts}. As the failures and serviceability are not interdependent in such networks, the connectivity probes do not relay any information regarding the failures; making it near impossible to infer the failures given the connectivity probes. 
The problem is compounded by the fact that the number of failure sets which do not effect the serviceability is exponential in terms of number of such edges. As we show in our experiments, using only connectivity probes returns failure set $I$ where recall is low as it fails to identify the edges which have no effect on the serviceability. 

Hence, a small sample of point probes of the failures may be helpful in boosting the power of connectivity probes as adding a sample of $I$ informs how many nodes to fail and may help to idenfity the true disaster scenario $o$.
Thus, additionally to $\Q_c$ we assume that a small set of (costlier) point queries $\qi \subseteq I$ (a sample of actual components that have failed) is also available.  
For the ease of modeling, we assume that edges $ e \in \qi$ are sampled uniformly at random with probability $\gi$~\cite{adhikari2018near}. In the rest of the paper, we refer to $\qi$ as the failure probe set and refer to both $\qi$ and $\qs$ together as the joint probes.

\subsection{MDL} 
Note that we do not know apriori how many failures are present in the system. Hence, intuitively, to formulate the problem above, we need a model selection framework which finds this automatically in a principled fashion. 
To this end, we propose to use the information-theoretic Minimum Description Length (MDL) principle~\cite{grunwald04tutorial}. 
We use two-part MDL, also known as the sender-receiver framework as we are specifically interested in the model. Our goal is to transmit the given set of probes $\mathcal{Q} = \qs \cup \qi$ (the `data') from sender to receiver in the least cost (in bits), by assuming that both of them know the layout of the network $G$, the sampling rates $\gs$ and $\gi$, the possible failure sources $\sourceSet$ and the conditional failure parobabilities $F's$.  
The MDL cost function consists of two parts: (a) Model cost includes the complexity of the selected model that explains the state of the network; and (b) Data cost that represents the cost of sending the given probe data $\mathcal{Q}$ given the model chosen. Formally, given a set of models $\mathcal{M}$, MDL claims the best model $M^*$ is the one that minimizes $\mathcal{L}(M)+ \mathcal{L}(\mathcal{D} \given M)$, in which $\mathcal{L}(M)$ is the model-cost (length in bits to describe model $M$), and $\mathcal{L}(\mathcal{D} \given M)$ is the data-cost (the length in bits to describe $\mathcal{D}$ using $M$). 

\subsection{Model Space and Cost}\label{subsec:sizeformmodelcost}
The first step in our formulation is to identify the right model \textit{space}.
Our failure process states that the failure set $I$ is induced due to a particular disaster scenario $o \in \mathcal{O}$. Hence, we include both $o$ and $I$ in the model. 
In real networks different disasters induce failures in totally different regions of the network. Hence, marginalizing over different scenarios and picking the most likely failure set $I$ over all disasters\cite{adhikari2018near} does not work well.

\begin{table}[t]
\centering

\caption{Table of symbols}
\begin{tabular}{|p{1.5cm}|p{5.5cm}|}
    \hline
    \textbf{Symbol} & \textbf{Definition} \\
    \hline
    $V_S, V_T, V_D$ & Supply, transshipment, and demand node set, respectively \\\hline
    $\mathcal{O}$ & Set of disaster scenarios \\\hline
    $I$& Set of failed edges (failure set) \\\hline
    $F(e|o)$& Failure probability of edge $e$ given disaster $o$ \\\hline
    $\s$ & Set of serviced nodes (serviced set) \\\hline
    $\gs, \gi$ & Probability for (uniformly) sampling probes from sets $\s$ and $I$, respectively \\\hline
    $\qs$ & Connectivity probe set (connectivity queries) sampled from set $\s$ \\\hline
    $\qi$ & Point probe set (point queries) sampled from set $I$ \\\hline
    $h(G, \s, I)$ & Indicates if the concurrence of $G$, $\s$, and $I$ is possible (returns 1) or not (returns 0) \\\hline
\end{tabular}
\label{tab:notation}
\end{table}

We want to primarily infer only $I$, but this set by itself is not enough to be the model because there is no direct relationship between $I$ and $Q_c$; thus, it would not be useful to explain the data. Therefore, we use all three of $o$, $I$ and $\s$ as our model: $\mathcal{M} = (o, \s, I)$. In other words, we first send the disaster scenario $o$, then send the true serviced set $\s$, followed by the true failed set $I$. We then identify the actual joint probes set $\qs$ and $\qi$ as the data.

\setlength\abovedisplayskip{1pt}
\setlength\belowdisplayskip{1pt}
The MDL model cost
$\mathcal{L}(o, \s, I)$ has the following three components: 
\begin{align*}
\mathcal{L}(o, \s, I) = \mathcal{L}(o) +\mathcal{L}(\s \given o) + \mathcal{L} ( I \given \s, o )
\end{align*}

Using the Shannon-Fano code we have the following.
{
\begin{align*}
    \mathcal{L}(o) &= - \log (Pr(o) ) \\
    \mathcal{L}(\s \given o) &= -\log ( Pr(\s \given o) ) \\
    \mathcal{L}(I \given \s, o) &= -\log ( Pr(I \given \s, o) )
\end{align*}

}
Combining and expanding these, we obtain the following.
\vspace{0.05in}
{ 
\begin{align*}
&\mathcal{L}(o, \s, I) \\
&=- \log (p(o)) -\log \big(Pr(\s \given o)\big) - \log \left( \frac{Pr(\s \given I, o) Pr (I \given o)}{Pr (\s \given o)} \right) \\
&=   - \log (p(o)) - \log \big(  Pr(\s \given I, o) Pr (I \given o) \big)
\end{align*}
}

\noindent
where $Pr(o) = p(o)$.  
Note from \eqref{eq:causal} that given $I$, $\s$  does not depend longer on $o$. Hence we have, $Pr( \s \given I, o) =  Pr(\s \given I)$. 
Clearly, $Pr(\s \given I)$ is 1 if after removing $I$ from $G$, the serviced set is exactly $\s$, and 0 otherwise.
For $Pr(I\given o)$, recall that we assume that edge failures depend only the disaster's effect and are independent from each other, therefore, we can represent $Pr(I\given o)$ as a product of failure probabilities of the individual failure. Note that $Pr(\s \given I)$  is the probability of set of demand nodes $S$  being  connected to the source the given failure set is $I$. As this probability can only take value of $0$ or $1$ (indicating whether it is feasible for $S$ to be connected or not), we represent it with a indicator function $h(G, \s, I)$, which we refer to as a feasibility function. 
Combining all of the above, the final model cost is:

{
\begin{align}
&\mathcal{L}(o, \s, I) = \nonumber \\
&- \log \big(p(o)\big)  -\log \big( h(G, \s, I)  \prod_{e\in I} F( e | o) \prod_{e\in E\setminus I} (1 - F( e | o) ) \big).
\label{eq:final_model_cost_cq}
\end{align}
}

\subsection{Data Cost} 
Note that MDL needs a data cost as well (otherwise we will choose a model without considering $\qi$ and $\qs$).
In the MDL framework, given the model is already transmitted, we now send the data. 
For the data cost, we send first the size of $\qs$ and $\qi$, which are $\norm{\qs}$ and $\norm{\qi}$, respectively. They followed by the joint queries, i.e. our two probes sets $\qs$ and $\qi$.
Hence, our data cost consists of four terms. 

As $\Q_c$ is sampled uniformly from $\s$ with probability $\gs$, the size $\norm{\qs}$ is send as follows:
\vspace{0.08in}
\setlength\abovedisplayskip{5pt}
\setlength\belowdisplayshortskip{5pt}
{ 
\begin{align*}
\mathcal{L} ( \norm{\Q_c} | \s, I, o )  &= - \log ( Pr(\norm{\Q_c} | \s, I, o)) \nonumber \\
&=  -\log ( \binom{\norm{\s}}{\norm{\Q_c}} \gs^{\norm{\Q_c}} (1 - \gs)^{\norm{\s \setminus \Q_c}})
\end{align*}
}
Similarly, for $\norm{\qi}$ we have: 
\setlength\abovedisplayskip{5pt}
\setlength\belowdisplayshortskip{5pt}
{ 
\begin{align*}
\mathcal{L} ( \norm{\qi} | \s, I, o )  &= - \log ( Pr(\norm{\qi} | \s, I, o)) \nonumber \\
&=  -\log ( \binom{\norm{I}}{\norm{\qi}} \gi^{\norm{\qi}} (1 - \gi)^{\norm{I \setminus \qi}})
\end{align*}
}

Then the  cost of  sending $\qs$ and $\qi$ in terms of previously sent information is: 
\vspace{0.08in}
\setlength\abovedisplayskip{5pt}
\setlength\belowdisplayshortskip{5pt}
{
\begin{align*}
&\mathcal{L}(\qs \given[\big] \s, I, \norm{\qs}) + 
\mathcal{L}(\qi \given[\big] \s, I, \norm{\qi}) \nonumber \\
= &- \log (Pr(\qs \given[\big] \s, I, \norm{\qs}) ) 
- \log (Pr(\qi \given[\big] \s, I, \norm{\qi}) ) \nonumber \\
= & - \log ( \gs^{\norm{\qs}} (1-\gs)^{\norm{\s \setminus \qs}} )
- \log ( \gi^{\norm{\qi}} (1-\gi)^{\norm{I \setminus \qi}} )
\end{align*}
}

Combining the four data cost terms, the total data cost is:
{
\begin{align}
\label{eq:datacost_joint}
= &- \log \binom{\norm{I}}{\norm{\qi}} - 2\norm{\qi} \log (\gi) - 2(\norm{I} - \norm{\qi}) \log(1 - \gi) \nonumber \\
& -\log \binom{\norm{\s}}{\norm{\Q_c}} - 2\norm{\Q_c} \log (\gamma_c) - 2(\norm{\s} - \norm{\Q_c}) \log(1 - \gamma_c).
\end{align}
}

\noindent\textbf{Formal Problem Statement.}
We can now state our problem formally as:
\begin{problem}
\label{prob:problem}
\prob{}: \textit{Given an undirected graph $G(V ,E)$ where $V=V_S \cup V_D \cup V_T$,
a many-to-many relation mapping supply nodes $V_S$ to demand nodes $V_D$,
a set of observed serviced nodes $\Q_c$ sampled uniformly at random with probability $\gamma_c$ from $\s$, 
and a set of observed failed edges $\qi$ sampled uniformly at random with probability $\gi$ from $I$,
find the complete set of serviced nodes $\s^* \subseteq V_D$ and failed edges $I^* \subseteq E$ (that failed as per the failure model described in Section~\ref{sec:prelim}),
by minimizing the MDL cost function,
i.e.} 
{\small 
\begin{align*}
&\langle \s^*, I^* \rangle \nonumber\\
&= \underset{\s, I}{\arg\min}  \Big\{ \nonumber\\
&- \log \big(p(o)\big)  -\log \big( h(G, \s, I)  \prod_{e\in I} F( e | o) \prod_{e\in E\setminus I} (1 - F( e | o) ) \big) \nonumber\\
&- \log \binom{\norm{I}}{\norm{\qi}} - 2\norm{\qi} \log (\gi) - 2(\norm{I} - \norm{\qi}) \log(1 - \gi) \nonumber \\
& -\log \binom{\norm{\s}}{\norm{\Q_c}} - 2\norm{\Q_c} \log (\gamma_c) - 2(\norm{\s} - \norm{\Q_c}) \log(1 - \gamma_c)
\Big\},
 \nonumber
\end{align*}
}
i.e.,
{\small 
\begin{align}
\label{eqn:objective}
&\langle \s^*, I^* \rangle =  \underset{\s, I}{\arg\min}  \Big\{ \mathrm{Eq.~(\ref{eq:final_model_cost_cq})} + \mathrm{Eq.~(\ref{eq:datacost_joint})} \Big\}
\end{align}
}
\end{problem}

\vspace{6pt}

\noindent\textbf{Algorithm for \prob{} Problem.}
Solving Problem \ref{prob:problem} is very challenging. First of all, the search space of the solution is exponentially large as it involves searching over all possible sets of serviced and failed nodes, and handle the combinatorial explosion of paths.
This makes the problem naturally more challenging than past work in point queries \cite{adhikari2018near}.
Moreover, since the objective is a function of two interdependent sets, designing any algorithm with performance guarantee is of a major challenge. 
Formally, we state the following lemma.

\begin{lemma}
\label{lemma:hardness}
\prob{} is NP-hard to approximate within an $\Omega(\sqrt{\log\log{|V|}})$ factor.
\end{lemma}
\vspace{5pt}
\textit{Proof.} (Sketch)
Our proof is by a reduction from the multicut problem, which is defined in the following manner: we are given an
instance $H=(V_H, E_H)$, and a set of source-sink pairs $(s_1, t_1),\ldots, (s_k, t_k)$. The goal is to pick 
the smallest subset $E'\subseteq E$ of edges, such that all the source-sink pairs are disconnected in $G[E\setminus E']$,
which was shown by \cite{Chawla2006} 
to be NP-hard to approximate within an $\Omega(\sqrt{\log\log{|V_H|}})$ factor,
under standard complexity theoretic assumptions.
We reduce this to an instance of \prob{} in the following manner.
We add a new supply node $s$, and add $m'$ parallel edges $(s, s_i)$ for each $i$ (this can be changed to non-parallel
edges, by adding a new node on each such edge). Finally, we add an additional demand node $t_{k+1}$, and add
the path from $s$ to $t_{k+1}$, which gives us the graph $G$.
We set $V_D=\{t_1,\ldots,t_{k+1}\}$, and $\s=\Q_c=\{t_{k+1}\}$, with $\gamma_c=1$.
Let $M$ be number of edges  in $G$. We set $F(e|o)=1/2^M$ for all edges $e$. 
Observe that any feasible solution must make all the terminals $\{t_1,\ldots,t_k\}$ disconnected from $s$.
Next, the cost of any feasible solution $I\subset E$ is $\leq M|I|+1$, because of the choice of $F(\cdot)$.
Finally, due to the parallel edges $(s, s_i)$, none of those edges will be picked in the solution.
This implies, the solution $I$ picked for this instance is also a multicut in $H$.
Since the cost of the solution remains the same, within a fixed multiplicative factor, the lemma follows.
$\blacksquare$
{
\begin{algorithm}[t!]{}
\caption{\jointalgolocal{}}
\begin{algorithmic}[1]
\small
    \STATE \textbf{Input:} Instance $(G, V, \qs, F, \gs, \qi, \gi)$ 
    \STATE \textbf{Output:} Solution $\hat{I}$ as per Eq. \eqref{eqn:objective} 
    \FOR{$o\in\sourceSet$}
    \STATE $\hat{I}(o) \gets \qi$ \quad \{initialize solution set for $o$\}
  \WHILE{$\exists e \in E \setminus \hat{I}(o)$ that decreases the MDL cost}
    \STATE \{To calculate MDL, first compute $\s$ with BFS in $G \setminus \hat{I}(o)$\}
    \STATE Select $e \in E \setminus \hat{I}(o)$ that decreases the most the MDL cost        
        \STATE $\hat{I}(o) \gets \hat{I}(o) \cup \{ e \}$ \quad \{update solution\}
        \STATE $\alpha(o) \gets $ MDL cost for updated $\hat{I}(o)$
    \ENDWHILE
    \ENDFOR
\STATE Return $\hat{I}(o)$ for $o$ that minimizes $\alpha(o)$
\end{algorithmic}
\label{alg:localsearch}
\end{algorithm}
}
\par \noindent 
\textbf{Note:} It is worth observing that the corresponding point query only formulation of \cite{adhikari2018near} can be approximated
within an additive $O(\log{n})$ factor. Additive approximations are, in general, easier than multiplicative
approximation guarantees. Therefore, Lemma \ref{lemma:hardness} implies Problem \ref{prob:problem} is a 
harder problem than the point query version, from a complexity standpoint.

\prob{} is quite challenging as it is related to the multicut problem with multiple paths between source-sink pairs. Since a related multicut problem leverages a greedy strategy \cite{kortsarts2005greedy}, we adopt this approach. We describe our algorithm \jointalgolocal{} for \prob{} next. 
It iterates over the disaster scenarios in $\mathcal{O}$ to find a failure set for each $o$.
To obtain $\hat{I}(o)$, it starts by initializing it to $\qi$. Then, at each iteration, an edge $e \in E\setminus \hat{I}(o)$, whose addition decreases the MDL cost the most, is added to the failure set $\hat{I}(o)$. 
The process is repeated until no such edge can be found. At the end, the $\hat{I}(o)$ that has the lowest MDL score is returned.
The implementation needs the calculation of $\s$ after $\hat{I}(o) \cup \{e\}$ is removed from $G$, which needs to check which demand nodes has at least one path to a supply node, 
which is accomplished via breadth-first search (BFS).
The complete pseudocode is in Algorithm~\ref{alg:localsearch}.
\newpage
\begin{lemma}
Algorithm \jointalgolocal{} runs in $O(|V|^2 |E|+|V||E|^2)$ time.
\end{lemma}
\textit{Proof.} Calculating total MDL cost for any solution instance $\hat{I}$ requires a BFS (to calculate the $\hat{\s}$ induced by $\hat{I}$) for each demand node, which, takes $O(|V|^2+|V||E|)$. The inner loop in lines 5-10 takes $O(|E|)$ and the loop in 3-11 takes $O(|\mathcal{O}|)$. In practice, $|\mathcal{O}| \lll |V|$ is a small constant; therefore, our algorithm runs in $O(|V|^2 |E|+|V||E|^2)$.
$\blacksquare$

Note that the feasibility function contained in the MDL cost will ensure that the infeasible solutions are not selected. This is because the term $h(G, \s, I) = 0$ for infeasible solutions. Hence, the MDL cost for such solutions will be infinite and they will not be picked by our heuristic. Hence, the algorithm always ensures that $\qs \subseteq \hat{\s}$.

\section{Experiments}
\label{sec:expts}
\noindent
In this section, we empirically test our formulations and algorithms in both single and multiple disaster scenarios.

\subsection{Setup}
\par \noindent
\textbf{Setup.}
We implemented the algorithms in Python\footnote{Code and data: \href{https://github.com/arodriguezca/JointPathMap}{https://github.com/arodriguezca/JointPathMap}}\!.
We ran our algorithms on a Linux machine with 8 cores with 3.40 GHz and 16GB of RAM. 
Presented results are the average over 30 different \emph{damage scenarios} simulated as per civil engineering standards (explained in detail in our supplement). Each damage scenario is originated from a single \emph{scenario earthquake} $o\in\sourceSet$ (each $o$ is a magnitude and an epicenter) which was randomly selected with probability $p(o)$.
\jointalgolocal takes roughly 2 seconds to output a solution (for a single damage scenario) for EPN and PWN networks, and roughly 500 seconds for our large network GridKit (see Section \ref{subsec:dataset} for network specifications). Therefore, our algorithm is fast and scalable.

\subsection{Datasets}
\label{subsec:dataset}
\par \noindent
We evaluate the performance of our algorithms on the topologies of three real infrastructure networks. For each network, we follow procedures as described in civil engineering literature~\cite{adachi_comparative_2010} with structural parameters from~\cite{fema_multi-hazard_2013} to generate multiple geographical correlated failure probabilities for nodes. Edge failure probabilities are calculated from these counterparts. 

\begin{figure}[h]
    \centering
    \includegraphics[width=0.44\textwidth]{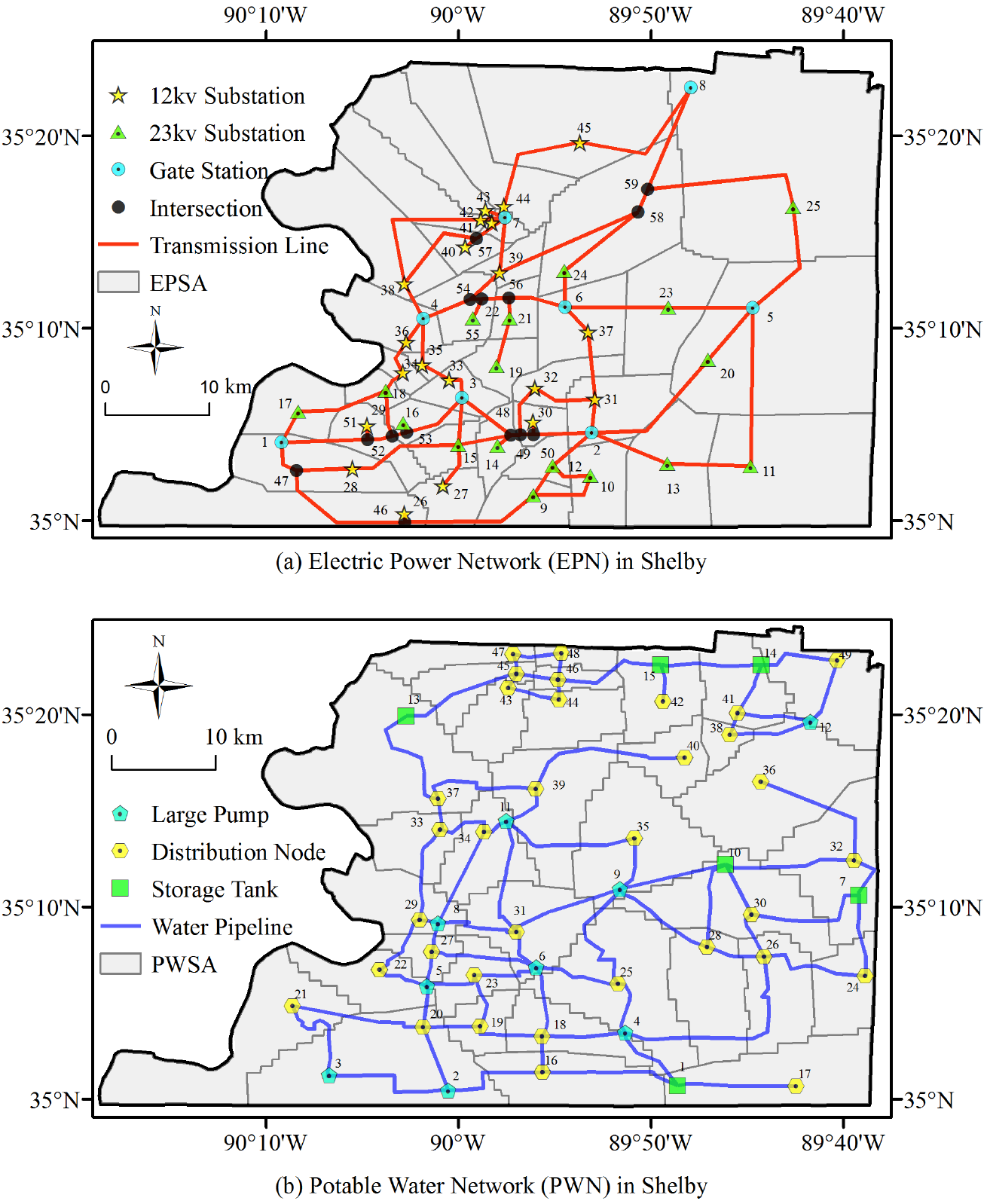} 
    \caption{Skeletonized power and water network topologies in Shelby County, Tennessee.} 
    \label{fig:shelby}
\end{figure}

\par \noindent
\textbf{EPN \& PWN.} 
We use two utility networks for Shelby County, TN, USA: electric power network (EPN) and potable water network (PWN), with $\norm{V_{EPN}}=59$, $\norm{E_{EPN}}=73$, $\norm{V_{PWN}}=49$, $\norm{E_{PWN}}=71$, whose topologies are depicted in Figure \ref{fig:shelby} 
and are publicly available in~\cite{zhang_probabilistic_2018-1}. 
We adopt the 10 scenario earthquakes that characterize the seismic hazard in the area identified in \cite{adachi_comparative_2010}. This means that once an earthquake happens, its effects on structures can be characterized by one of these 10 scenario earthquakes.
Each of the scenario earthquake induce different failure probability-sets ($F$'s) for the network.

\par \noindent
\textbf{GridKit.}
We use the high-voltage power grid for Northern California, obtained from \cite{wiegmans2016gridkit}, from which we extract the subgraph corresponding to Northern California.
It contains $\norm{V_{GK}}=415$ and $\norm{E_{GK}}=543$. We generate 50 scenario earthquakes, each of them as follows: epicenter and magnitude are selected randomly, the first from all node locations, 
and the second from the four values considered in \cite{adachi_comparative_2010}.

Table~\ref{tab:network_stats} contains the specific number of demand, source and transshipment nodes for these three networks. As mentioned before, \jointalgolocal is fast in these real networks publicly available; therefore, when larger networks become available, our algorithm will apt to run in a suitable time.

\begin{table}[h]
\centering
\caption{Datasets details}
\begin{tabular}{ m{1cm} | m{1cm} | m{1cm} | m{1cm} | m{1cm} | m{1cm} } 
Network & $\norm{V}$ & $\norm{V_S}$ & $\norm{V_D}$ & $\norm{V_T}$ & $\norm{E}$\\ \hline 
EPN & 59 & 9 & 37 & 13 &  73\\
PWN & 49 & 15 & 34 & 0 & 71 \\
GridKit & 415 & 114 & 82 & 219 & 543
\end{tabular}
\label{tab:network_stats}
\end{table}

We consider the following factual observation in our experiments. 
Even though the network topology allows it, in practice, source nodes cannot service \emph{any} demand node, but only the ones close to them that its generative capacity allows. To model this physical constraint, we restrict serviceability of a demand node to the case when there is at least one path of length $L$ to any source. In our experiments, we use $L_{EPN}=3$, $L_{PWN}=2$, and $L_{GK}=4$.

\subsection{Baselines and Metrics}
We use two baselines, namely \modelcost and \pathmap. \modelcost is  an algorithm that greedily minimizes model cost in Eq.~\eqref{eq:final_model_cost_cq}, to demonstrate the importance of considering the data cost. \pathmap is a baseline that only includes connectivity queries.

To evaluate our algorithms, we use 
F1-score, for which we need to compute precision and recall. Following the notation in our paper, let $I$ be the true failure set and $\hat{I}$ be our solution set. Our metrics are defined in literature as follows: $\text{precision} = \norm{\hat{I} \cap I } / \norm{\hat{I}}$, 
$\text{recall} = \norm{\hat{I}  \cap I } / \norm{I}$, and  $\text{F1-score} = \text{precision} \times \text{recall} / (\text{precision} + \text{recall})$. We add the following limit cases to handle empty $I$ or $\hat{I}$: precision is 1 when $\hat{I}=I=\emptyset$ (we correctly identified no failures), and recall is 1 when $\norm{I}=0$ (there are no failures missed in our solution $\hat{I}$). 

\subsection{Results}
\begin{figure}[t]
\centering
\includegraphics[width=0.33\textwidth]{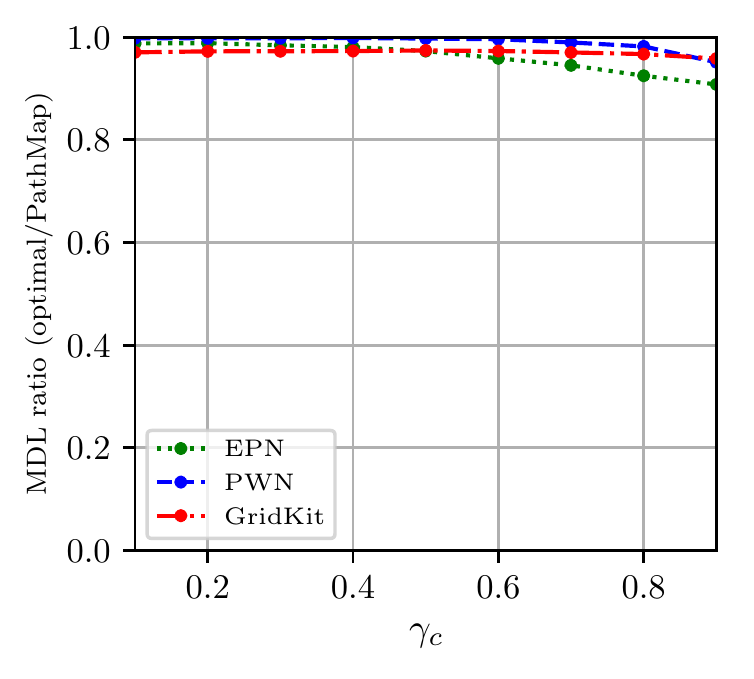}
\linespread{1}
\caption{MDL ratio between optimal and our method \jointalgolocal. }
\label{fig:mdl_u_edges}
\vspace{-0.1in}
\end{figure}

\prob{} problem asks to optimize the MDL cost to infer the complete failure set, and \jointalgolocal{} succeeds at this optimization task at low computation cost.
This is reflected in our results in Figure~\ref{fig:mdl_u_edges}. We can see that the ratio between optimal MDL and our methods is close to 1. 
The optimal MDL is calculated via integer linear programming where all paths are listed.

\begin{figure*}
    \centering
    \includegraphics[width=0.9\textwidth]{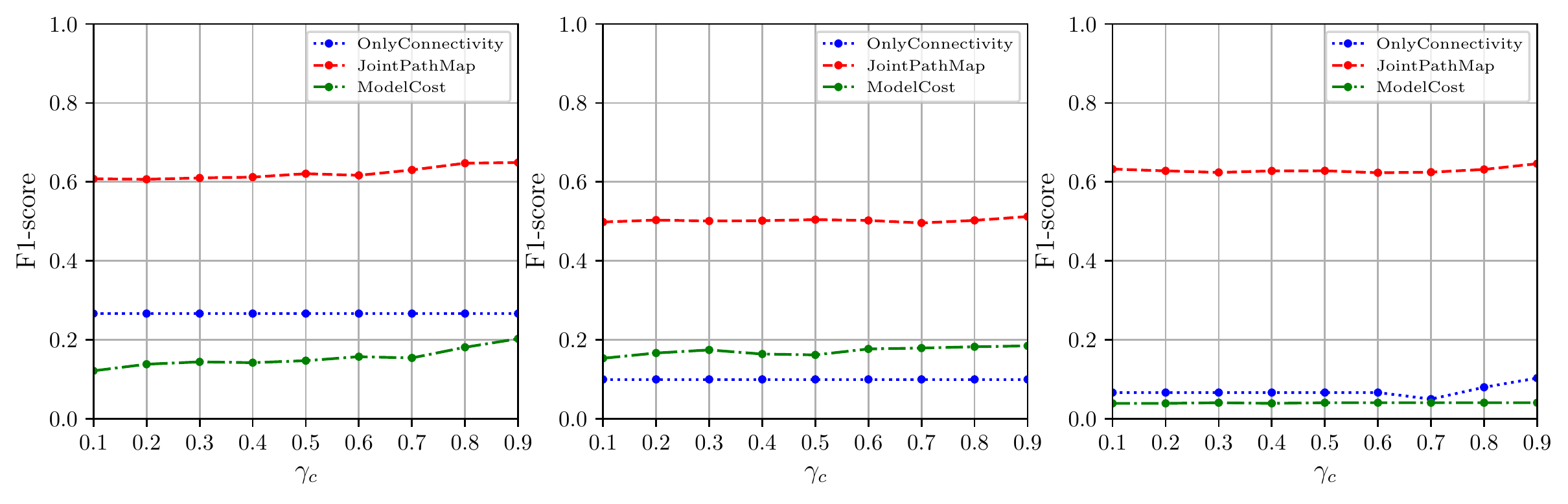}
    \caption{F1-score for  \jointalgolocal ($\gamma_I=0.3$), \algolocal and \modelcost for multiple disaster scenarios. Datasets from left to right: EPN, PWN, and GridKit.}
    \label{fig:results}
\end{figure*}

We present the performance of all the methods in Figure \ref{fig:results} in terms of F-1 score. 
As described before, this score is an average over 30 damage scenarios, each of them coming from a different disaster scenarios $o$ sampled from $\sourceSet$ (our scenario earthquakes) with probabilities $p(o)$. As failure probabilities $F$ depend on $o$, this set of damage scenarios leads to a variety of ground truth failure sets $I$. 
As we can observe, \jointalgolocal outperforms all the baselines significantly in all three real world networks for all values of $\gs$. Poor performance of \modelcost can be attributed to the fact that it infers the most likely failures while ignoring the observed data. 
Thus, \modelcost's result highlights the necessity of the MDL framework.

\begin{figure}
\centering
\includegraphics[width=0.33\textwidth]{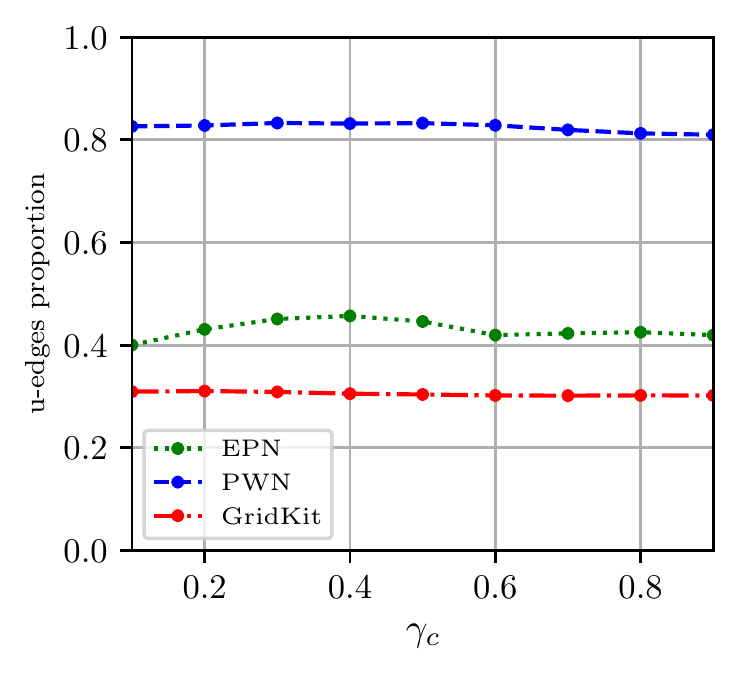}
\caption{
u-edges proportion is an estimate of the proportion of existing edges in the actual failure set $I$ that are not reflected in the connectivity queries. This proportion is especially high for the PWN, which is at the same time the hardest to solve for \pathmap.
}
\label{fig:u_edges}
\end{figure}

Similarly, \pathmap has low F1-score due to low recall, which indicates that while the inferred failures are mostly correct, they are just a fraction of the total failure set. This is due to the fact that, depending on the redundancy of the network topology, some of the failures are not reflected in the connectivity queries, for which it would be very hard to infer by only using $\qs$. To get a sense of how many of 'undiscoverable' edges (u-edges) are present in ground truth failure set $I$, we add as many edges from $I$ to $\hat{I}$ as long serviceability is not affected (i.e. $\hat{\s}$ does not change). Then, we calculate the proportion of u-edges in $I$.
In Figure~\ref{fig:u_edges}, we can see that there is a high proportion of u-edges, i.e. edges not reflected in connectivity queries. This indicates there are many failure sets that can produce the same serviced set $\s$; consequently, solely using connectivity probes $\qs$ is clearly not sufficient. 
As shown by the results, \jointalgolocal does not suffer from this issue and works well in real datasets. The improved performance is because \jointalgolocal incorporates a small set of probes of the failures $\qi$, which provides insight about the location of the epicenter thanks to the geographical correlation among failures.

\begin{figure}
\centering
\includegraphics[width=0.33\textwidth]{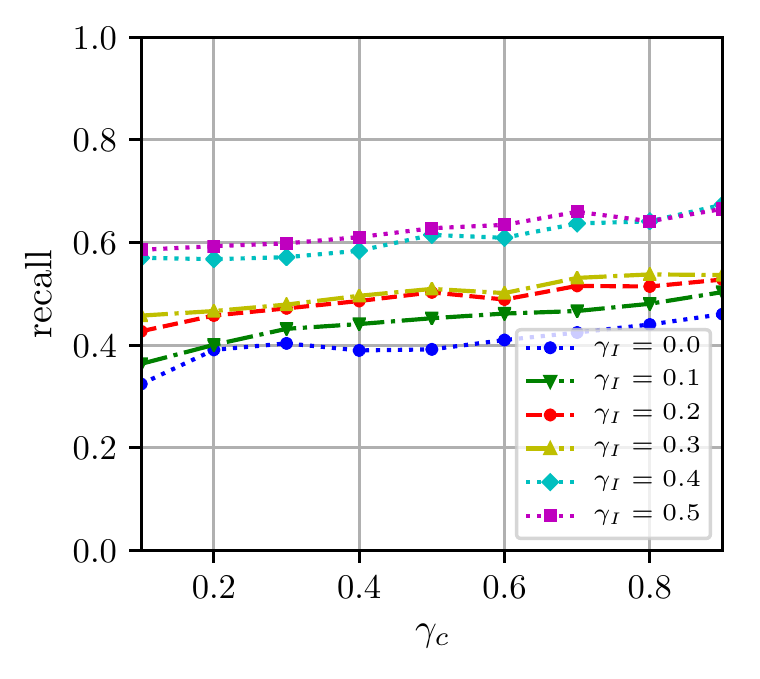}  
\caption{The effect of increasing $\gamma_I$ (i.e. increase the sample of failure probes) for EPN dataset. 
We present recall here because we want to emphasize the ability of our algorithm to discover elements of the failure set $I$ that are not in $\qi$.}
\label{fig:mdl_u_edges_joint}
\end{figure}

For multiple scenario earthquakes, the algorithm has to infer first the culprit scenario $o$ and then the complete failure set $I$.  
In a small network, some of scenario earthquakes are similar and missing the right one may not drastically affect further inference; however, in large networks, scenarios earthquakes are completely different (mainly in epicenter), therefore, without a proper inference of $o$ it is very hard to make a good inference of $I$. 
Adding a small sample of failure probes helps to properly identify $o$ and the right expected number of failures, and gives a larger improvement in F1-score. 
In addition, we can see in Figure \ref{fig:mdl_u_edges_joint} the effect of different sample sizes $\gamma_I$, where each increment has a different effect, being the most relevant the jump from $\gi=0.3$ to $\gi=0.4$.

\begin{figure*}
    \centering
    \includegraphics[width=0.9\textwidth]{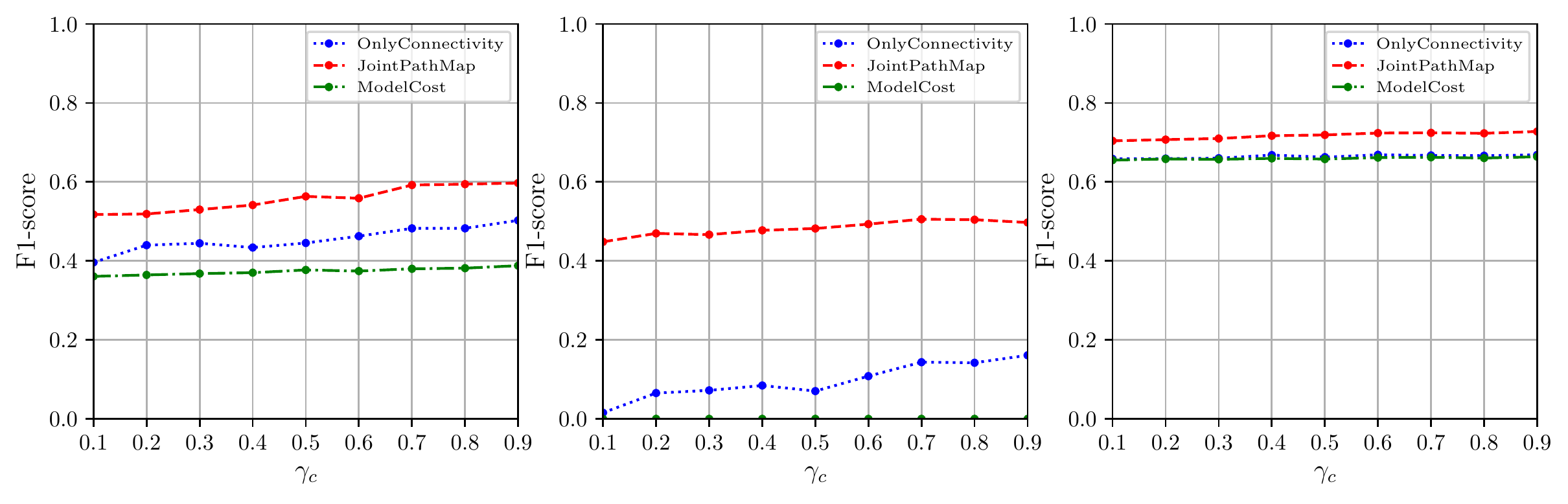}
    \caption{F1-score for \jointalgolocal ($\gamma_I=0.3$), \algolocal and \modelcost for a single disaster scenario, i.e. $o$ is known. Datasets from left to right: EPN, PWN, and GridKit. \modelcost returns empty solutions ($\hat{I}=\emptyset$) for PWN because all F's are below 0.5 (i.e. are more likely to remain functional), but the actual expected failure set size is $\approx 10$. Therefore, as its recall is 0, its F1-score is constantly 0.}
    \label{fig:results_single}
\end{figure*}

As additional results, we also present results for a single disaster scenario. 
In practice, after an earthquake occurs, an estimate of its magnitude and epicenter is released within minutes, but this information is then corrected several times over the next hours. Therefore, for this kind of natural disaster, it is preferrable to handle uncertainty by considering multiple earthquake scenarios as we did in our formulation. However, once the estimation about magnitude/epicenter is reliable, we can add this extra information to our problem by reducing $\sourceSet$ to a single $o$ and making $p(o)=1$. 
For our single disaster scenario experiments, we pick the $o$ that has the largest expected number of failures for EPN and PWN; and, for GridKit, as most disaster scenarios lead to similar damage, we select one at random. 
As expected, when know the disaster scenario $o$, this extra information helps the inference as we can see in Figure \ref{fig:results_single}. As we consider the scenario which has the most failures for EPN and PWN (the hardest one), the performance of \jointalgolocal may be slightly lower than when aggregating the performance over all scenarios. 

\begin{figure}
\centering
\resizebox{0.48\textwidth}{!}{\input{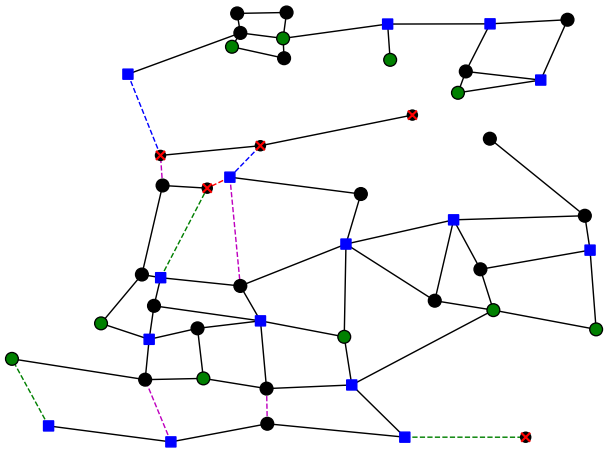}}
\caption{Visualization of $I$, $\s$ and $\hat{I}$ in PWN for $\gamma_c=\gamma_I=0.3$. 
Source nodes are depicted as squares and demand nodes as circles (there are no transshipment nodes).
Probes $\qs$ and $\qi$ are represented as green nodes and edges respectively.
Ground truth non-serviced demand nodes have a red X.
Ground truth failed edges are represented as dotted lines. 
Blue edges are predicted failures by \jointalgolocal.
Red edges are discoverable failures missed by \jointalgolocal.
Purple edges are u-edges.
}
\label{fig:net_vis}
\end{figure}

\subsection{Case Study}
Within the single scenario earthquake framework, we study how the failures discovered relate to the ground truth failure set in PWN. 
In Figure \ref{fig:net_vis}, we observe that our algorithm misses four failures, from which three of them are actually undiscoverable (u-edges). In other words, these failures do not impede any demand node (circle) to be serviced by source nodes (squares) as it can be recognized from the figure. Serviceability in this network is restricted to available paths of length 2 within source and demand node. Consequently, in this network, there are five demand nodes non-serviced (marked with a red X), for which our algorithms found the right failures to disconnect for four of them. In fact, the two failed edges found (dashed blue edges) are the reason why three demand nodes are non-serviced. Therefore, repairing them is of utmost importance to recover service in most of the non-serviced demand nodes. 
This exemplifies how \jointalgolocal finds the truly important informative edges to restore serviceability.

\section{Related Work}
\noindent 
\par \noindent \textbf{Critical Infrastructure Networks:} Data mining critical infrastructure has been gaining much research interest in recent times~\cite{tabassum2018data}. These include works in modeling~\cite{ouyang2014review}, dependency inference~\cite{chen2016fascinate}, and finding hotspots~\cite{chen2017hotspots}.
Chen et al.~\cite{chen2016fascinate} proposed collective collaborative filtering based approach for inferring dependency between various interconnected critical infrastructure networks. Similarly Chen et al.~\cite{chen2017hotspots} proposed cascade based failure model in heterogeneous critical infrastructure networks to identify most critical nodes in the system. 
A more closely related line of work studies failure inference on infrastructure networks. \cite{anwar2015roadrank} proposed an algorithm to infer congested intersections given the road networks and usage data. Similarly \cite{adhikari2018near} proposed a near-optimal MDL-based approach to infer failures, but given point probes. 

\noindent\textbf{Network Tomography:} This involves inferring the network states based on end-to-end network measurements. These measurements, available in the Internet and other communication networks, 
measure delay across signals transmitted by the probes, which is information that can be used to infer delays on links~\cite{xia2006inference} and network topology~\cite{jin2006network,anandkumar2013topology}. 
Xia and Tse~\cite{xia2006inference} propose an expectation maximization algorithm to infer delays on links from performing a great quantity of end-to-end measurements.
Jin et al.~\cite{jin2006network} proposed a heuristic to infer approximate topology based on end-to-end measurements with the goal of reducing the number of such measurements while maintaining accuracy.
Anandkumar et al.~\cite{anandkumar2013topology} study inferring the topology of sparse networks. 
They evaluate their algorithm with two routing models and find that they only require a sub‐linear number of participant nodes selected uniformly at random. 

\noindent\textbf{Epidemic Inference:} 
Another closely related field is reverse engineering an epidemic outbreak over a network given partially observed infections. Shah et al proposed a MLE based approach to detect the source of computer virus spreading over a computer network~\cite{shah2010detecting}. Similarly, \cite{prakash2012spotting,sundareisan2015hidden} leverage MDL based approach to detect both the source and missing infections in networks. These works assume that the infection/virus propagation over the network takes a stochastic form often modelled using a variant of the popular SIR model. On the other hand, a different set of works \cite{rozenshtein2016reconstructing, xiao2018reconstructing, xiao2018robust} model the interaction among individuals as a temporal or static time expanded network and treat the combination of the observed infections and the time as a node in the network. The inference is typically done by solving temporally constrained Steiner-tree problem where the terminals are the observed infections.

\section{Conclusions}
\label{sec:conc}

\noindent We investigated the problem of inferring network states given connectivity queries, a piece of information more practical to obtain than a direct sample of network states (point queries).
Our empirical observations indicate that redundancies, common in infrastructure networks, cause a significant proportion of states to be undiscoverable by connectivity probes. Therefore, we propose adding to the connectivity queries a small sample of point queries to exploit relationships among network states, which proved to be valuable for boosting the power of connectivity queries. We formulated the novel problem with MDL and developed a scalable greedy algorithm that consistently optimizes MDL cost close to optimal in extensive experiments. These experiments are performed in real networks and principled simulations. 
Future work can focus on extending to node failures and using noisy probe data. Other interesting future work is extending our formulation to other applications of connectivity probes such as LAN networks and industrial communication networks.

\section*{Acknowledgements}
This paper is based on work partially supported by the NSF (Expeditions CCF-1918770, CAREER IIS-2028586, RAPID IIS-2027862, Medium IIS-1955883, NRT DGE-1545362), CDC MInD program, ORNL, funds/computing resources from Georgia Tech, and the National Institute of Standards and Technology through the Center for Risk-Based Community Resilience Planning [70NANB15H044, 70NANB20H008]. 
A.V.'s work was partially supported by NSF grants CRISP 2.0 Grant 1832587, CMMI-1745207, IIS-1633028 and DTRA subcontract/ARA S-D00189-15-TO-01-UVA.

\newpage


\bibliographystyle{IEEEtran}

\providecommand{\noopsort}[1]{}

\section{Appendix}

\subsection{Data generation}
\label{sec:dg}
In this paper, we used simulation data developed as per civil engineering standards that encapsulates structural and soil considerations.
To generate the different earthquake scenarios for the networks introduced in our paper (EPN and PWN), we followed the procedure described in \cite{adachi2009serviceability,adachi_comparative_2010}. 
First, we used the attenuation equation (Eq.~\eqref{eq:attenuation}) of \cite{toro_model_1997} to determine the median peak ground acceleration (PGA) on each location, which is associated to the seismic magnitude $M_w$ and epicentral distance $R$.
\setlength\abovedisplayskip{5pt}
\setlength\belowdisplayskip{5pt}
\begin{multline}
\ln(\mbox{median PGA}) = 2.2 + 0.81(M_w-6.0) \\
- 1.27 \ln(\sqrt{R^2+9.3^2}) - 0.0021 \sqrt{R^2+9.3^2} \\
+0.11 \, \max \left[\ln\left(\frac{R}{100}\right), 0.0 \right]
\label{eq:attenuation}
\end{multline}
In particular, we estimated the PGA at the location of each component in the network using simulated seismic events 
consistent with \cite{adachi2009serviceability,adachi_comparative_2010}. Then, in order to determine the failure probabilities associated with each component, we used the fragility curves described in \cite{fema_multi-hazard_2013}.  

\begin{table}[ht!]
\centering
\caption{Median PGA at which the components of the studied networks reach extensive damage, thus, considered failed. Intersection nodes in EPN are considered invulnerable.}
\vspace{3mm}
\label{tab:param}
\begin{tabular}{llrr}
\toprule
Network & Component Type  & Median (g) & $\beta_{sd}$\\ 
\midrule
EPN   & Gate station        & 0.47        & 0.40 \\
        & 23-kV substation    & 0.70       & 0.40  \\
        & 12-kV substation    & 0.90       & 0.45 \\ 
\midrule
PWN   & Storage tanks & 0.68  &  0.75  \\
        & Pumping stations    &  0.80   & 0.80 \\ 
        & Distribution nodes    &  1.15  & 0.60 \\ 
\end{tabular}
\end{table}

In \cite{fema_multi-hazard_2013}, it is assumed that given $S_d$ --- a spectral displacement or other relevant Potential Earth Science Hazard (PESH) parameter --- the probability associated with being in or exceeding a given damage state $ds$ can be modeled using a cumulative lognormal distribution, as described in Eq. \eqref{eq:probFail}. In Eq. \eqref{eq:probFail}, $\bar{S}_{d,ds}$ is the median value of spectral displacement or PESH parameter at which the building or component of interest reaches the damage state $ds$, $\beta_{sd}$ is the standard deviation of the natural logarithm of spectral displacement or PESH parameter for damage state $ds$, and $\Phi$ is the standard normal cumulative distribution function \cite{fema_multi-hazard_2013},
\setlength\abovedisplayskip{5pt}
\setlength\belowdisplayskip{5pt}
\begin{eqnarray}
P[ ds|S_d ]=\Phi \left[ \frac{1}{\beta_{sd}} \text{ln} \left( \frac{S_d}{\bar{S}_{d,ds}} \right) \right].
\label{eq:probFail}
\end{eqnarray}

For our particular case study, the PESH used is the PGA \cite{fema_multi-hazard_2013} at the location of each component of the EPN and PWN, calculated with Eq. \eqref{eq:attenuation}. The values of $\beta_{sd}$ and the median PGA at which the components of the studied networks reach a damage state $ds$ consistent with complete or extensive damage (depending on the component type) are described in Table \ref{tab:param} \cite{adachi2009serviceability,adachi_comparative_2010,fema_multi-hazard_2013}. We consider a component in this damage state as failed. Note that these components are nodes in our networks.

After calculating the probability of failure of each node, we calculate the edge failure probabilities. We assume the failure probability of an edge depends on its incident nodes (if one of the nodes fails, the edge fails). Therefore, $F(e\given o ) = F(n_1\given o ) + F(n_2\given o ) - F(n_1\given o )F(n_2\given o )$, where $e=(n_1,n_2)$. Finally, having these failure probabilities, we used Monte-Carlo simulation to generate the 30 earthquake-consistent damage scenarios used in this paper.

\end{document}